\pdfoutput=1

\documentclass[11pt]{article}
\usepackage[dvipsnames]{xcolor}

\usepackage[preprint]{acl}

\usepackage{times}
\usepackage{latexsym}
\usepackage{url}
\usepackage[T1]{fontenc}

\usepackage[utf8]{inputenc}

\usepackage{microtype}

\usepackage{inconsolata}
\usepackage[dvipsnames]{xcolor}

\usepackage{graphicx}
\usepackage{multirow}

\usepackage{booktabs}
\usepackage{amsmath}

\usepackage{pgf}
\usepackage{colortbl}

\newcommand{\gradientcell}[6]{%
    \def\value{#1}%
    \def\minvalue{#2}%
    \def\maxvalue{#3}%
    \def\mincolor{#4}%
    \def\maxcolor{#5}%
    \def\transparency{#6}%
    \ifdimcomp{\value pt}{>}{\maxvalue pt}{\cellcolor{#5!100.0!#4!#6}\value}{%
    \ifdimcomp{\value pt}{<}{\minvalue pt}{\cellcolor{#5!0.0!#4!#6}\value}{%
         \pgfmathparse{int(round(100*(#1/(\maxvalue-\minvalue))-(\minvalue *(100/(\maxvalue-\minvalue)))))}%
        \xdef\tempa{\pgfmathresult}%
        \cellcolor{#5!\tempa!#4!#6}\value%
    }}%
}

\definecolor{LightGray}{gray}{0.95} 
\newcommand{\negcorr}[1]{\gradientcell{#1}{-1.0}{0.0}{MidnightBlue}{LightGray}{50}}
\newcommand{\poscorr}[1]{\gradientcell{#1}{0.}{1.0}{LightGray}{BrickRed}{50}}
\newcommand{\C}[1]{%
  \ifdim#1pt<0pt
    \negcorr{#1}%
  \else
    \poscorr{#1}%
  \fi
}
\newcommand{\R}[1]{\gradientcell{#1}{0.18}{4.85}{LightGray}{ForestGreen}{50}}

\title{Evaluating Morphological Plausibility of Subword Tokenization via Statistical Alignment with Morpho-Syntactic Features}

\author{Abishek Stephen \and Jindřich Libovický \\
  Charles University, Faculty of Mathematics and Physics \\
  Institute of Formal and Applied Linguistics \\
  Malostranské náměstí 25, 118 00 Prague, Czech Republic \\  
  \texttt{\{stephen, libovicky\}@ufal.mff.cuni.cz}}
  
\begin{document}
\maketitle
\begin{abstract}

We present a novel metric for the evaluation of the morphological plausibility of subword segmentation.
Unlike the typically used morpheme boundary or retrieval F-score, which requires gold segmentation data that is either unavailable or of inconsistent quality across many languages, our approach utilizes morpho-syntactic features.
These are available in resources such as Universal Dependencies or UniMorph for a much wider range of languages.
The metric works by probabilistically aligning subwords with morphological features through an IBM Model~1.
Our experiments show that the metric correlates well with traditional morpheme boundary recall while being more broadly applicable across languages with different morphological systems.
%

\end{abstract}

\section{Introduction}

Subword tokenization is a fundamental preprocessing step in modern NLP systems. When evaluating tokenizers, researchers consider both extrinsic metrics (downstream task performance) and intrinsic properties, including morphological plausibility—how well tokenization aligns with morphological segmentation \citep{uzan-etal-2024-greed,libovicky-helcl-2024-lexically,arnett-bergen-2025-language}—alongside statistical measures like compression ratio and vocabulary coverage \citep{limisiewicz-etal-2023-tokenization,zouhar-etal-2023-tokenization,schmidt-etal-2024-tokenization}.

Evaluating morphological plausibility faces challenges due to unavailable or inconsistent gold standard segmentation data, which biases experiments toward high-resource languages. Even among these languages, cross-dataset inconsistencies exist. For example, the German word \textit{Wolkenkratzer} ("skyscraper") is segmented as \textit{Wolke + n + kratz + er} in German-CELEX \citep{gulikers1995german} but left unsegmented in German MorphyNet \citep{batsuren-etal-2021-morphynet}, raising concerns about evaluation validity.

\begin{figure}[h]
\includegraphics[width=\columnwidth]{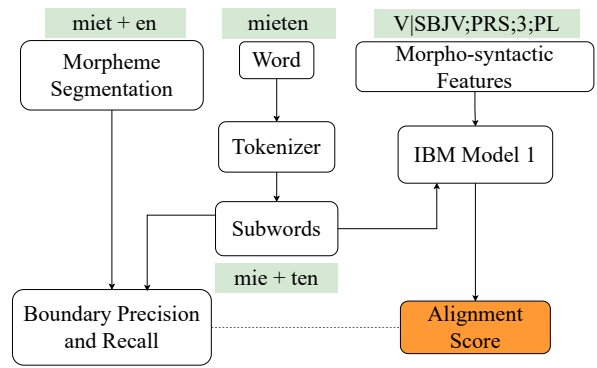}
\caption{%
The workflow for subword tokenization evaluation for morphological plausibility, illustrated with an example of the German verb \setlength{\fboxsep}{1pt}\colorbox[HTML]{d5e8d4}{\textit{mieten}} (``to rent''): Instead of comparing subwords with morpheme segmentation, we align morpho-syntactic features with subwords using IBM Model~1 to compute our proposed \setlength{\fboxsep}{0pt}\colorbox[HTML]{FF9933}{alignment score}.
}

\label{fig:workflow}
\end{figure}

Additionally, some tokenizers employ non-concatenative approaches, using latent variables \citep{samuel-ovrelid-2023-tokenization}, incomplete text coverage \citep{hofmann-etal-2022-embarrassingly}, or additional tags for casing and diacritics \citep{popel-etal-2022-cuni,tokenmonster,semenov2025inca}. For these approaches, traditional morpheme boundary metrics are not well-defined.

We address these limitations with a novel metric that evaluates morphological plausibility without gold segmentation data, instead utilizing morpho-syntactic features from resources such as UniMorph \citep{batsuren-etal-2022-unimorph}, available for 169 languages. These features enable cross-lingual comparison through a language-independent schema, such as the Dutch word \textit{adviseren} (``to advise'') tagged as \texttt{V;SBJV;PRS;PL} (Verb; Subjunctive mood, Present tense, Plural).

Our method uses IBM Model~1 \citep{brown-etal-1993-mathematics} to align morphological features with subword tokens, making it applicable to both concatenative and non-concatenative tokenization schemes. We extract alignment probabilities between subword tokens and morpho-syntactic features and aggregate them into a single measure. The metric shows a strong correlation with traditional boundary recall across diverse languages.
The alignment-based approach rests on the principle that morphemes and morphological features are inherently linked. Well-segmented subwords should capture units that consistently express particular grammatical functions, leading to strong feature-subword alignments. Poor segmentation produces arbitrary character sequences that align weakly with any specific features, resulting in lower scores.

We conceptualize the workflow as illustrated in Figure~\ref{fig:workflow}. The data curation block involves collecting datasets with morpheme segmentation and morphosyntactic features mapped to word forms. We train tokenizers and the IBM Model~1 using the created data structure (Table~\ref{tab:data}) and propose an alignment score that quantifies the morphological plausibility. The datasets, along with the experimental codes\footnote{\url{https://github.com/abishekjs/morph-tok-eval}}, are also released. 

\begin{table}[t]
\footnotesize

\begin{tabular}{ccc}
\toprule
\textbf{Word} & \textbf{Segmentation} & \textbf{Morpho-syntactic Feature}      \\ \midrule
rýžový        & rýž|ov|ý              & \tt ADJ;ACC;MASC;INAN;SG  \\
bázeň         & báz|eň                & \tt N;ACC;SG;FEM          \\
projet        & pro|je|t              & \tt V;V.PTCP;MASC;PASS;SG \\ \bottomrule

\end{tabular}
\caption{The curated dataset by combining the segmentation from Universal Segmentations and morpho-syntactic features from UniMorph. }
\label{tab:data}

\end{table}

\section{Related Work}

\subsection{Tokenizer Evaluation}

Most frequently used subword tokenizers are trained with a statistical heuristic, such as greedily shortening the training corpus \citep{sennrich-etal-2016-neural} or minimizing negative log likelihood of the training data in a unigram model \citep{kudo-2018-subword}.
The properties of the resulting tokenizer depend not only on the algorithm itself but also, to a large extent, on data preprocessing and the languages in the training data mix. 

Intrinsic evaluation of tokenizers includes information-theoretical properties, such as compression ratio or Rényi efficiency \citep{zouhar-etal-2023-tokenization}. In multilingual setups, the evaluation can include vocabulary allocation for different languages \citep{limisiewicz-etal-2023-tokenization} or literal and semantic token overlap between languages \citep{hammerl-etal-2025-beyond}.

Morphological qualities of word segmentation are usually evaluated either via morpheme precision and recall in more linguistic contexts \citep{batsuren-etal-2022-sigmorphon} or via morpheme boundary precision and recall in the context of subword segmentation \citep{uzan-etal-2024-greed,libovicky-helcl-2024-lexically}.


\subsection{IBM Models}

IBM Models are a family of unsupervised statistical models originally developed for word alignment \citep{brown-etal-1993-mathematics} that was a crucial component in statistical machine translation. IBM Model~1, the simplest in this family, learns lexical translation probabilities $P(f \mid e)$ that word $e$ translates as $f$. It is trained using an Expectation-Maximization algorithm on parallel data. In the E-step, the model computes expected alignment counts based on current probability estimates (i.e., the frequency with which words occur aligned to each other); in the M-step, it re-estimates translation probabilities from these counts.

IBM Models for word alignment in statistical machine translation have previously been shown to discover the relationship between morphemes and morphosyntactic features \citep{stephen-2024} and can be used for the unsupervised extraction of morphological categories for morphemes. These results suggest that the alignment probabilities may be a reliable indicator of the morphological quality of subword segmentation.

\section{The Alignment Score}

Our metric uses IBM Model~1 to establish probabilistic alignments between subword tokens and morphological features. It operates as an expectation-maximization algorithm that learns translation probabilities between source and target elements by iteratively maximizing the likelihood of observing the target given the source, without requiring any initial alignment. In our context, it discovers the probability distribution $P(f \mid s)$ between subword $s$ tokens and morphological features $f$ by treating each subword-feature pair as a potential alignment, then converging toward alignments that best explain the observed co-occurrences in the data.

The morphological plausibility score for tokenizer $T$
\begin{equation}
\frac{1}{|W|} \sum_{w \in W} \frac{1}{|S_w|} \sum_{s \in S_w} \operatorname{agg}_{f \in F_w} P(f \mid s)
\end{equation}
where $W$ is our corpus, $F_w$ are features for word $w$, $S_w$
are subwords of $w$, $\operatorname{agg}$ is a function that aggregates the probability scores for a single subword.
We implement various aggregation functions, including maximum, minimum, sum of probabilities, sum of logarithms, and mean. 

To eliminate noisy alignments, we only consider probabilities $P(f\mid s)$ over a certain threshold, which is a hyperparameter of our method.

\begin{table*}[ht]
\newcommand{\lnghead}[1]{\begin{minipage}{5mm}\centering #1\end{minipage}}

\footnotesize
\setlength{\tabcolsep}{2.8pt}
\begin{tabular}{cl rrrrrrrrrr@{\hskip 10pt} rrrrrrrrrr}
\toprule
\textbf{Type} &
  \multirow{2}{*}{\begin{minipage}{8mm}\vspace{-.8pt}\bfseries Ag\-gre\-gate\end{minipage}} &
  \multicolumn{10}{c}{\textbf{Boundary Precision}} &
  \multicolumn{10}{c}{\textbf{Boundary Recall}}\\
\cmidrule(lr){3-12} \cmidrule(lr){13-22}
\multicolumn{2}{c}{} &
  \lnghead{cs} &
  \lnghead{de} &
  \lnghead{en} &
  \lnghead{fi} &
  \lnghead{hr} &
  \lnghead{hu} &
  \lnghead{hy} &
  \lnghead{kn} &
  \lnghead{nl} &
  \lnghead{sk} &

  \lnghead{cs} &
  \lnghead{de} &
  \lnghead{en} &
  \lnghead{fi} &
  \lnghead{hr} &
  \lnghead{hu} &
  \lnghead{hy} &
  \lnghead{kn} &
  \lnghead{nl} &
  \lnghead{sk}  \\
\midrule
\multirow{5}{*}{Joint} & Sum & \C{ .06} & \C{-.28} & \C{-.67} & \C{-.14} & \C{-.13} & \C{.71} & \C{.32} & \C{-.65} & \C{-.24} & \C{.30} &  \C{ .94} & \C{.84} & \C{.67} & \C{.78} & \C{.74} & \C{-.04} & \C{.73}  & \C{.91} & \C{.80} & \C{.71} \\

& Log  & \C{ .27} & \C{ .72} &  \C{.79}& \C{-.15} & \C{-.35} & \C{.56} & \C{-.08} & \C{-.78} & \C{.81} & \C{.16}&\C{-.67} &\C{-.84}  &  \C{-.78} & \C{-.69} & \C{-.29} & \C{-.38} & \C{.49} & \C{.99} & \C{-.95} &  \C{.86}                 \\

& Mean & \C{ .05} & \C{-.35} &  \C{-.70}& \C{-.17} & \C{-.13} & \C{.60} & \C{.32} & \C{-.65} & \C{-.24} & \C{-.32} & \C{ .94}  & \C{.87}  &  \C{.71} & \C{.82} & \C{.74} & \C{.11} & \C{.73} & \C{.91} & \C{.80} &   \C{.70}                   \\

& Min  & \C{ .04} & \C{-.36}  & \C{-.70} & \C{-.15} & \C{-.13} & \C{.42} & \C{.32} & \C{-.65} & \C{-.24} & \C{-.32}&  \C{ .94}  & \C{.88}  &  \C{.71} & \C{.95} & \C{.74} & \C{.30} & \C{.73} & \C{.91} & \C{.80} & \C{.70}                     \\

& Max  & \C{ .05} & \C{-.35}  &  \C{-.70}& \C{-.13} & \C{-.13} & \C{.61} & \C{.32} & \C{-.65} & \C{-.24} & \C{-.33}& \C{ .94}  & \C{.87}  &  \C{.71} & \C{.79} & \C{.74} & \C{.11} & \C{.73} & \C{.91} & \C{.80} & \C{.70}         \\ \midrule

\multirow{5}{*}{Split} & Sum  & \C{-.03} & \C{-.31} & \C{-.70} &  \C{-.15}& \C{.02} & \C{.60} & \C{.15} & \C{-.67} & \C{-.17} & \C{-.35} & \C{ .98}  & \C{.87}  &  \C{.71} & \C{.82} & \C{.60} & \C{.11} & \C{.81} & \C{.95} & \C{.72}    & \C{.71}  \\
& Log  & \C{ .21} & \C{.60}  & \C{.79} & \C{.09} & \C{-.55} & \C{.65} & \C{-.29} & \C{-.71} & \C{.15} & \C{.44} & \C{-.96} & \C{-.88} &  \C{-.78} & \C{-.84} & \C{-.27} & \C{-.63} & \C{-.64}  & \C{.91} & \C{-.60} & \C{-.84}                     \\

& Mean & \C{-.04} & \C{-.34}  & \C{-.72} & \C{-.11} & \C{-.16} & \C{.58} & \C{.12} & \C{-.67} & \C{-.16} & \C{-.34} & \C{ .98}  & \C{.88} &  \C{.72} & \C{.91} & \C{.81} & \C{.15} & \C{.83} & \C{.96} & \C{.72} &         \C{.72}     \\

& Min  & \C{-.05} & \C{-.36} &  \C{-.72}& \C{-.02} & \C{-.00} & \C{.43} &\C{-.05} & \C{-.63} & \C{-.14} & \C{-.33} & \C{ .98}  & \C{.89} &  \C{.72} & \C{.94} & \C{.82}  & \C{.27} & \C{.88} & \C{.90} &  \C{.70}  &           \C{.73}    \\

& Max  & \C{-.04} & \C{-.35} &  \C{-.71}& \C{-.10} & \C{-.19} & \C{.60} & \C{.14} & \C{-.66} & \C{-.16} & \C{-.34}& \C{ .98}  & \C{.88} &  \C{.71} & \C{.80} & \C{.76} & \C{.11} & \C{.82} & \C{.97} &  \C{.71}   &            \C{72}  \\ \bottomrule
\end{tabular}
\caption{The tokenizer-level Spearman correlation of our metric scores across all tokenizers using different aggregates over full and split tags across languages. The threshold for the alignment probabilities obtained through IBM Model~1 is 0.01.}
\label{tab:results}
\end{table*}

\begin{figure*}[ht]
\includegraphics[width=\textwidth]{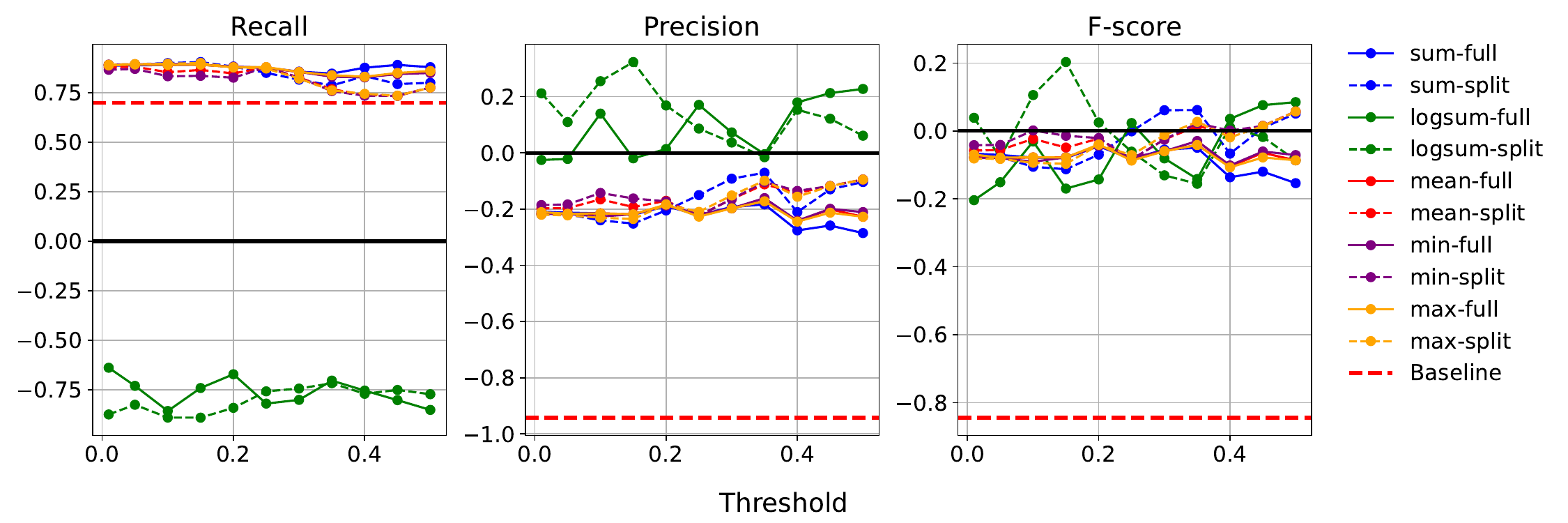}
\caption{The Spearman correlation of our metric scores with boundary recall, precision, and F$_1$ score for Finnish. Plots for other languages are in the Appendix.}
\label{fig:finnish}
\end{figure*}

\section{Experiments}

We validate the proposed metric by measuring the Spearman correlation between the proposed metric and the established method of measuring morphological plausibility, specifically morpheme boundary precision, recall, and F$_1$ score.

\subsection{Linguistic Resources}

We use segmentation resources from Universal Segmentations \cite{zabokrtsky-etal-2022-towards}, a data resource that houses harmonized datasets for morphological segmentation. The data resources used are Armenian-MorphyNet (hy), Finnish-MorphyNet (fi), Kannada-KCIS (kn), English-CELEX (en), German-CELEX (de), Dutch-CELEX (nl), Czech-DeriNet (cs), Serbo-Croatian-MorphyNet (hr), and Hungarian-MorphyNet (hu). Additionally, we use the Slovak (sk) dataset by \citet{olovstiak2015retrogradny}.

We use UniMorph \cite{batsuren-etal-2022-unimorph} for extracting the morpho-syntactic feature tags for the word forms. 

We test two approaches to feature representation: ``Joint'' tags containing complete morphological information, and ``Split'' tags where composite features are separated into atomic units. In ``Joint'' mode, a feature bundle like \texttt{ADJ;ACC;MASC;INAN;SG} is treated as a single token for IBM Model 1 alignment. In ``Split'' mode, each feature (POS, case, gender, animacy, number) becomes a separate token. Although considering morpho-syntactic features jointly makes more sense linguistically, treating them independently provides a richer training signal for estimating probabilities with the IBM model, which also helps with data sparsity.

The data for training the tokenizers and the metric was created by mapping the word forms along with their morphological segmentation with UniMorph feature tags.

\subsection{Subword Tokenizers}

We evaluate several standard subword tokenizers: Byte-Pair-Encoding \citep{sennrich-etal-2016-neural}, WordPiece \citep{schuster-nakajima-2012-japanese}, and the Unigram model \citep{kudo-2018-subword}. We train the tokenizers using 1M sentences from the CC100 corpus \citep{wenzek-etal-2020-ccnet} for the respective languages with vocabulary sizes of 2k, 4k, 8k, 16k, 24k, 32k, 40k, 48k, 56k, 64k, 72k, and 80k. Additionally, we add character segmentation and gold morphological segmentation to the correlation study.

We segment the UniMorph vocabulary using the tokenizers and run the IBM model for 10 epochs, which is enough for convergence. We evaluate the metric with 11 thresholds ranging from $0.01$ to $0.5$. Each tokenizer produced its own subword segmentation, which we then evaluate against the morphological features.

\subsection{Results}

We measure the correlation between the proposed morphological plausibility metric and traditional boundary-based metrics across 10 languages with different morphological structures. Table~\ref{tab:results} presents the Spearman correlation coefficients between our alignment-based scores and boundary precision/recall measures.

We present the results of the individual tokenizers in the Appendix in Table~\ref{tab:tokenizer}. Unigram tokenizers generally achieve better alignment scores than BPE and WordPiece, with smaller vocabularies typically yielding better results.

\paragraph{Correlation with Boundary Metrics.}

Our alignment-based metric correlates strongly with traditional boundary recall measures across the evaluated languages. Most languages show high positive correlations ($>0.70$) when using Sum, Mean, Min, and Max aggregation functions (See Appendix~\ref{sec:appendix}). Czech (cs) exhibits particularly strong correlations, reaching $0.94$-$0.98$ for boundary recall. Correlations with boundary precision vary more, with several negative correlations observed, except for Hungarian (hu), suggesting our metric aligns more closely with the recall aspect of morphological segmentation, rather than precision, which measures over-segmentation for almost all the languages. This is due to the arrangement in the training data, which is dominated by three feature tags, creating a strong co-occurrence pattern that makes it easier for the tokenizer to identify the correct morpheme boundaries.

\paragraph{Impact of Feature Representation.}

Split representation generally yields stronger correlations with boundary recall, particularly for morphologically complex languages such as Finnish (fi) and Armenian (hy). For Finnish (compare Table~\ref{tab:results} and Figure~\ref{fig:finnish}), the correlation increases from $0.79$ (Joint-Mean) to $0.91$ (Split-Mean). For the Hungarian dataset, we find stronger correlations with boundary precision, as the tokenizer finds correct boundaries due to the frequent morphological categories in the training data.

\paragraph{Aggregation Function Analysis.}

The Sum, Mean, Min, and Max aggregation functions exhibit similar performance patterns, with strong positive correlations with boundary recall. The Log aggregation function often yields inverse correlations (e.g., $-0.96$ for Czech, $-0.88$ for German). The Mean aggregation function performs consistently across languages, with an average correlation of $0.86$ with boundary recall when using split features.

\paragraph{Cross-lingual Observations.}

Performance varies across language families. The three Germanic languages (German, English, and Dutch) exhibit consistent correlation patterns, while Finnish and Kannada show some of the strongest correlations with the Min aggregation function ($0.94$ and $0.90$ respectively). Czech and Slovak correlate well with boundary recall but show more variable relationships with precision.

\section{Discussion}

The metric's inherent tendency to penalize over-segmentation can be seen through the weak correlation with the boundary precision scores.
However, we observe consistent patterns of correlations across languages, which provides a strong signal for the metric's cross-lingual viability. The metric is also able to capture the distinct morphology of languages well. Split representations exhibit higher correlations with boundary recall (especially for the Mean and Max functions) in Armenian, Czech, Finnish, Kannada, Serbo-Croatian, and Slovak, illustrating that the metric is sensitive to agglutination and allomorphy, with Hungarian being the only exception. This is additionally supported by almost identical results for Joint and Split representations for English and German, where we find a weaker fusional morphology. 

The performance of the metric is constrained by the granularity of morpheme segmentation and the alignment between morphemes and morpho-syntactic features. Since most morphemes align to features on a continuous rather than binary scale (even with thresholds), feature assignment lacks clear boundaries, reducing precision.

\section{Conclusion}

In this paper, we propose a new metric to assess the morphological plausibility of subword segmentation, addressing limitations of traditional evaluation metrics. We use datasets where the morpho-syntactic features of the word forms are also mapped, along with their morpheme segmentation. The morpho-syntactic features are taken from UniMorph, and the gold morpheme segmentation is accessed from Universal Segmentations. We train our tokenizers with varying vocabulary sizes using Byte-Pair-Encoding, WordPiece, and the Unigram model. The resulting subword tokens per word form are, respectively, aligned with the morpho-syntactic features using the IBM Model~1. 
The resulting alignment probabilities are aggregated in multiple ways to obtain a statistically robust array of results. These results are correlated with the traditional boundary precision, recall, and $F_1$ score obtained by comparing the tokenizer outputs with the gold morpheme segmentation data.

Our proposed metric correlates strongly with boundary recall across languages with varied morphological systems, making it a capable contender for testing morphological plausibility. 








\section*{Limitations}

The current stream of experiments is correlated with the existing gold segmentation data, which could potentially carry some inherent flaws. We do not run any checks whatsoever to quantify the accuracy of the gold segmentation data. 

\section*{Acknowledgments}

This research was supported by the Czech Science Foundation Project 25-16242S, Charles University student project GA UK No.\ 101924 and partially supported by SVV project number 260 698.

\bibliography{anthology,custom}

\appendix

\section{Appendix: Detailed Results}
\label{sec:appendix}

\begin{figure*}[]
\includegraphics[width=\textwidth]{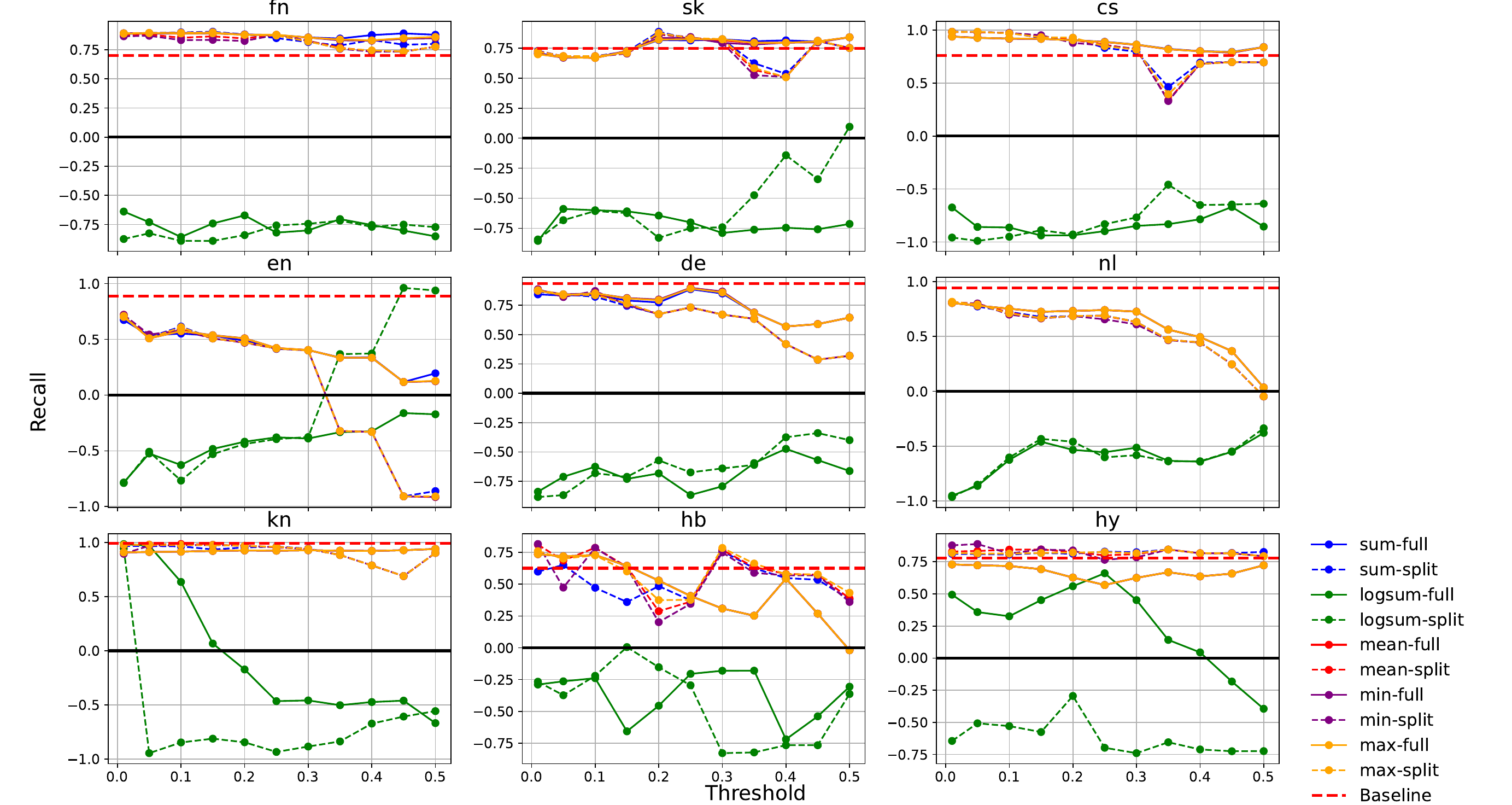}
\caption{The correlation of the alignment-based score with boundary recall for all languages.}
\label{fig:all_recall}
\end{figure*}

\begin{figure*}[]
\includegraphics[width=\textwidth]{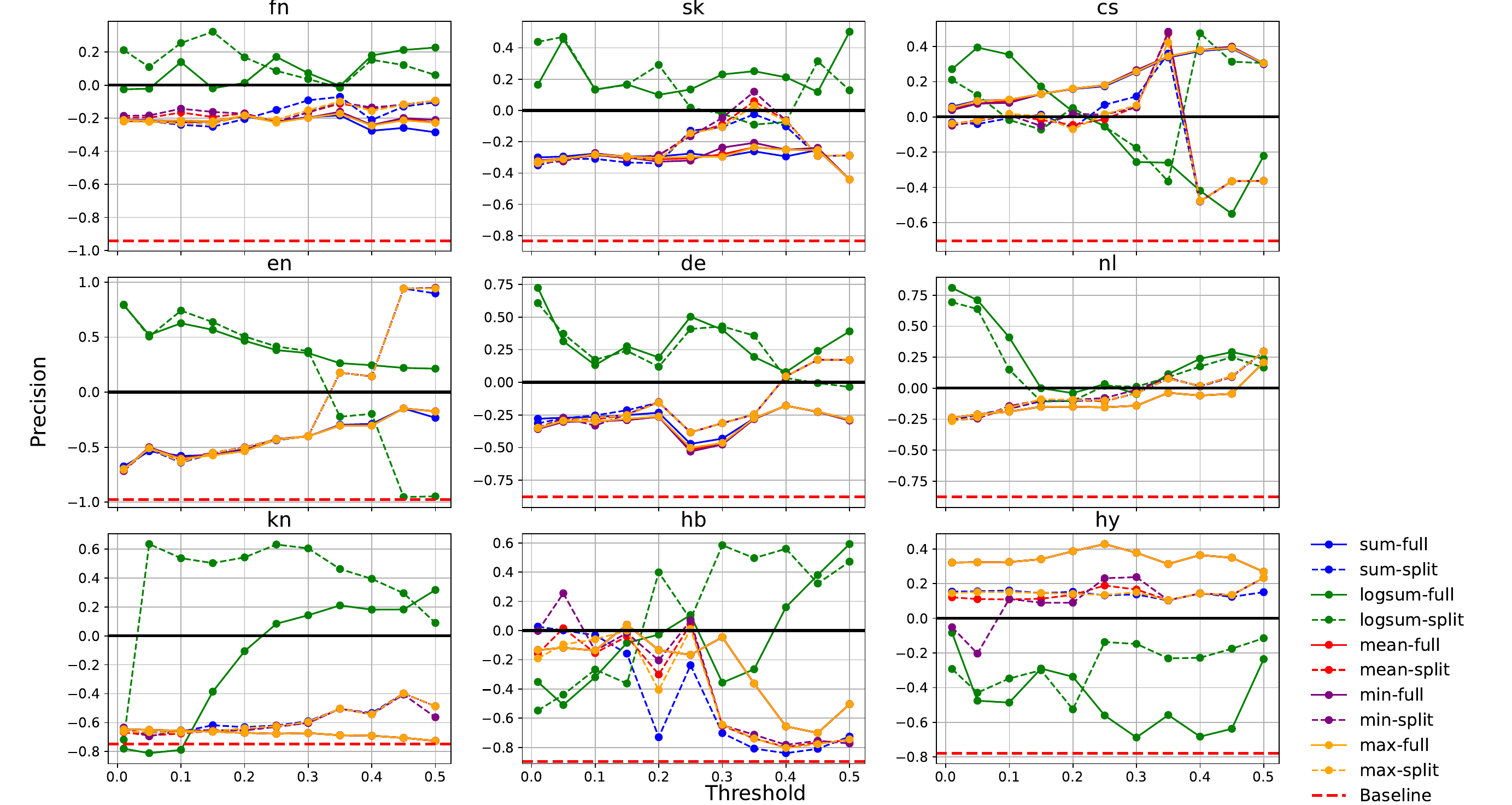}
\caption{The correlation of the alignment-based score with boundary precision for all languages.}
\label{fig:all_precision}
\end{figure*}

\begin{figure*}[]
\includegraphics[width=\textwidth]{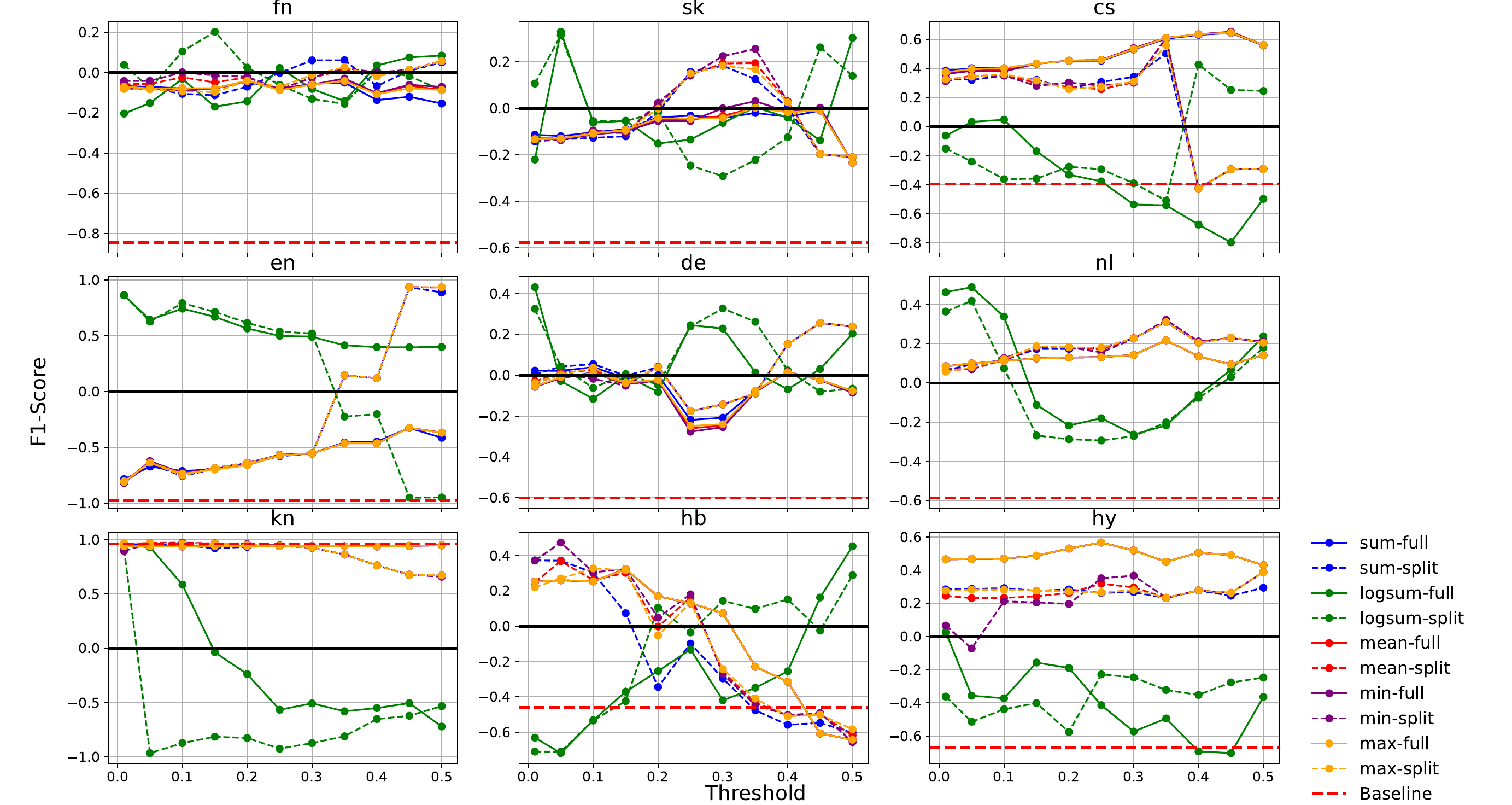}
\caption{The correlation of the alignment-based score with F$_1$ score for all languages.}
\label{fig:all_f1}
\end{figure*}

\begin{figure*}[]
\includegraphics[width=\textwidth]{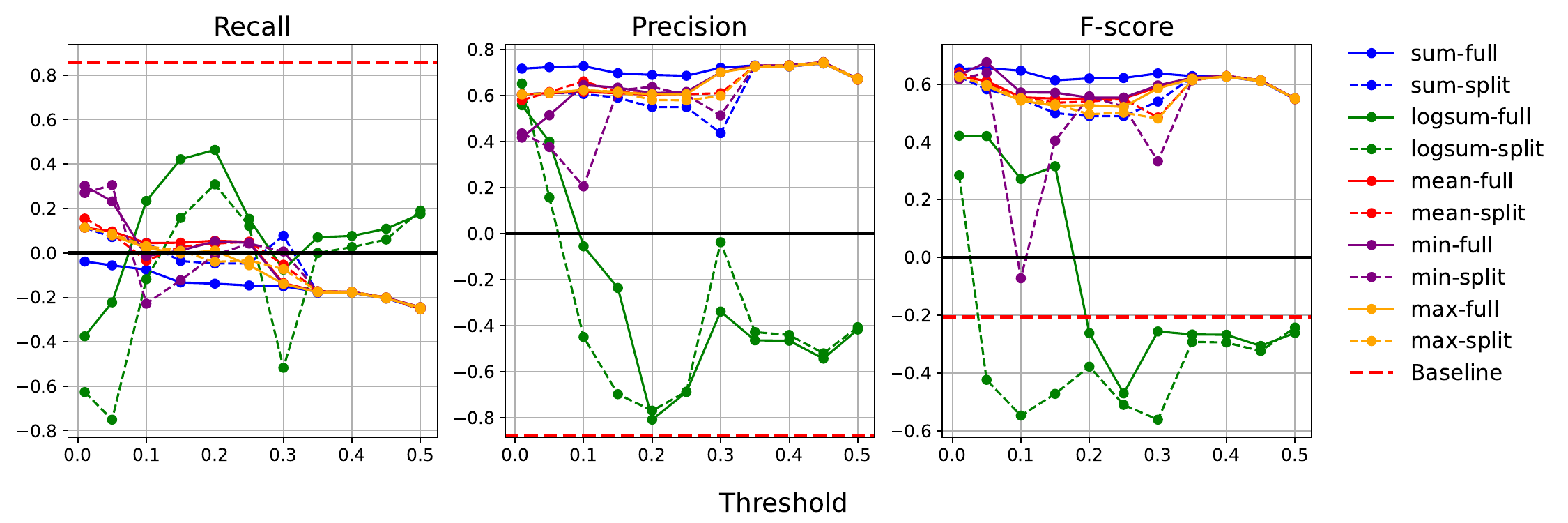}
\caption{The Spearman correlation of our metric scores with boundary recall, precision, and F$_1$ score for Hungarian.}
\label{fig:hun}
\end{figure*}

Figure~\ref{fig:all_recall} shows the correlation of the alignment score with boundary recall for different thresholds for 9 languages, Figure~\ref{fig:all_precision} shows the correlation with boundary recall, and Figure~\ref{fig:all_f1} shows the correlation with F$_1$ score. We present the results for Hungarian in Figure~\ref{fig:hun}.
Table~\ref{tab:tokenizer} shows the results of the individual tokenizers.

\begin{table*}
\footnotesize
\centering

\begin{tabular}{lr cccccccccc cc}
\toprule
Tokenizer & Vocab &
cs & de & en & fi & hu & hr & hy & kn & nl & sk &
\multicolumn{2}{c}{Average} \\
\midrule

\multirow{11}{*}{BPE}
&  4k & \R{0.67} & \R{0.77} & \R{0.65} & \R{2.04} & \R{1.46} & \R{1.55} & \R{2.09} & \R{2.59} & \R{1.21} & \R{1.44} & \R{1.45} & \multirow{11}{*}{\R{0.98}} \\
&  8k & \R{0.55} & \R{0.54} & \R{0.45} & \R{1.55} & \R{1.44} & \R{1.51} & \R{1.69} & \R{2.38} & \R{0.94} & \R{1.03} & \R{1.21} \\
& 16k & \R{0.40} & \R{0.41} & \R{0.35} & \R{1.23} & \R{1.50} & \R{1.49} & \R{1.61} & \R{2.10} & \R{0.71} & \R{0.80} & \R{1.06} \\
& 24k & \R{0.33} & \R{0.35} & \R{0.32} & \R{1.11} & \R{1.50} & \R{1.46} & \R{1.59} & \R{1.85} & \R{0.57} & \R{0.72} & \R{0.98} \\
& 32k & \R{0.29} & \R{0.32} & \R{0.31} & \R{1.03} & \R{1.51} & \R{1.47} & \R{1.56} & \R{1.78} & \R{0.46} & \R{0.65} & \R{0.94} \\
& 40k & \R{0.25} & \R{0.28} & \R{0.29} & \R{0.99} & \R{1.54} & \R{1.47} & \R{1.54} & \R{1.68} & \R{0.39} & \R{0.60} & \R{0.90} \\
& 48k & \R{0.23} & \R{0.26} & \R{0.28} & \R{0.93} & \R{1.56} & \R{1.48} & \R{1.59} & \R{1.60} & \R{0.32} & \R{0.58} & \R{0.88} \\
& 56k & \R{0.21} & \R{0.24} & \R{0.27} & \R{0.88} & \R{1.57} & \R{1.48} & \R{1.56} & \R{1.57} & \R{0.28} & \R{0.54} & \R{0.86} \\
& 64k & \R{0.20} & \R{0.22} & \R{0.27} & \R{0.85} & \R{1.58} & \R{1.43} & \R{1.54} & \R{1.58} & \R{0.25} & \R{0.51} & \R{0.84} \\
& 72k & \R{0.18} & \R{0.21} & \R{0.26} & \R{0.82} & \R{1.58} & \R{1.46} & \R{1.51} & \R{1.58} & \R{0.23} & \R{0.48} & \R{0.83} \\
& 80k & \R{0.18} & \R{0.19} & \R{0.26} & \R{0.78} & \R{1.57} & \R{1.45} & \R{1.49} & \R{1.56} & \R{0.21} & \R{0.46} & \R{0.82} \\
\midrule

\multirow{11}{*}{Unigram}
&  4k & \R{2.26} & \R{2.26} & \R{3.77} & \R{3.39} & \R{1.94} & \R{1.90} & \R{2.49} & \R{2.45} & \R{4.85} & \R{3.47} & \R{2.88} & \multirow{11}{*}{\R{2.21}} \\
&  8k & \R{2.03} & \R{1.80} & \R{2.85} & \R{2.43} & \R{1.76} & \R{1.93} & \R{2.22} & \R{2.67} & \R{4.20} & \R{3.17} & \R{2.51} \\
& 16k & \R{1.82} & \R{1.40} & \R{2.04} & \R{2.42} & \R{1.71} & \R{1.89} & \R{2.20} & \R{2.47} & \R{3.60} & \R{2.97} & \R{2.25} \\
& 24k & \R{1.71} & \R{1.22} & \R{1.69} & \R{2.46} & \R{1.75} & \R{1.91} & \R{2.19} & \R{2.30} & \R{3.27} & \R{2.91} & \R{2.14} \\
& 32k & \R{1.64} & \R{1.09} & \R{1.48} & \R{2.58} & \R{1.76} & \R{1.93} & \R{2.26} & \R{2.24} & \R{3.08} & \R{2.93} & \R{2.10} \\
& 40k & \R{1.56} & \R{1.02} & \R{1.37} & \R{2.69} & \R{1.76} & \R{2.00} & \R{2.41} & \R{2.27} & \R{2.92} & \R{2.94} & \R{2.09} \\
& 48k & \R{1.51} & \R{0.99} & \R{1.30} & \R{2.84} & \R{1.76} & \R{2.02} & \R{2.35} & \R{2.24} & \R{2.77} & \R{2.94} & \R{2.07} \\
& 56k & \R{1.47} & \R{0.97} & \R{1.24} & \R{2.94} & \R{1.77} & \R{1.95} & \R{2.36} & \R{2.21} & \R{2.57} & \R{3.00} & \R{2.05} \\
& 64k & \R{1.44} & \R{0.97} & \R{1.21} & \R{3.08} & \R{1.79} & \R{1.90} & \R{2.39} & \R{2.26} & \R{2.47} & \R{3.03} & \R{2.05} \\
& 72k & \R{1.42} & \R{0.93} & \R{1.18} & \R{3.22} & \R{1.79} & \R{1.89} & \R{2.39} & \R{2.27} & \R{2.38} & \R{3.12} & \R{2.06} \\
& 80k & \R{1.43} & \R{0.90} & \R{1.16} & \R{3.35} & \R{1.81} & \R{1.81} & \R{2.39} & \R{2.30} & \R{2.31} & \R{3.16} & \R{2.06} \\

\midrule
\multirow{11}{*}{WordPiece}
&  4k & \R{0.78} & \R{1.00} & \R{1.02} & \R{1.71} & \R{1.42} & \R{1.54} & \R{2.16} & \R{2.11} & \R{1.70} & \R{1.62} & \R{1.51} & \multirow{11}{*}{\R{1.00}} \\
&  8k & \R{0.72} & \R{0.71} & \R{0.60} & \R{1.52} & \R{1.05} & \R{1.50} & \R{1.48} & \R{2.33} & \R{1.42} & \R{1.23} & \R{1.26} \\
& 16k & \R{0.55} & \R{0.47} & \R{0.46} & \R{1.26} & \R{0.93} & \R{1.44} & \R{1.32} & \R{2.28} & \R{1.19} & \R{1.09} & \R{1.10} \\
& 24k & \R{0.50} & \R{0.40} & \R{0.40} & \R{1.13} & \R{0.90} & \R{1.44} & \R{1.27} & \R{2.06} & \R{1.01} & \R{1.06} & \R{1.02} \\
& 32k & \R{0.47} & \R{0.33} & \R{0.38} & \R{1.10} & \R{0.84} & \R{1.42} & \R{1.25} & \R{1.82} & \R{0.88} & \R{1.00} & \R{0.95} \\
& 40k & \R{0.45} & \R{0.30} & \R{0.37} & \R{1.06} & \R{0.86} & \R{1.44} & \R{1.26} & \R{1.69} & \R{0.77} & \R{0.95} & \R{0.92} \\
& 48k & \R{0.42} & \R{0.29} & \R{0.37} & \R{1.08} & \R{0.86} & \R{1.47} & \R{1.26} & \R{1.58} & \R{0.74} & \R{0.93} & \R{0.90} \\
& 56k & \R{0.39} & \R{0.26} & \R{0.35} & \R{1.05} & \R{0.86} & \R{1.47} & \R{1.27} & \R{1.58} & \R{0.67} & \R{0.88} & \R{0.88} \\
& 64k & \R{0.37} & \R{0.24} & \R{0.35} & \R{1.05} & \R{0.87} & \R{1.48} & \R{1.27} & \R{1.53} & \R{0.58} & \R{0.86} & \R{0.86} \\
& 72k & \R{0.36} & \R{0.21} & \R{0.35} & \R{1.01} & \R{0.86} & \R{1.49} & \R{1.25} & \R{1.50} & \R{0.52} & \R{0.82} & \R{0.84} \\
& 80k & \R{0.35} & \R{0.20} & \R{0.34} & \R{1.01} & \R{0.85} & \R{1.44} & \R{1.32} & \R{1.49} & \R{0.55} & \R{0.80} & \R{0.83} \\

\bottomrule

\end{tabular}

\caption{Alignment score $\times$ 100 for individual tokenizers and languages.}
\label{tab:tokenizer}
\end{table*}

\end{document}